\documentclass{article}

\usepackage[final]{corl_2022}

\usepackage{amsmath}
\usepackage{amssymb}
\usepackage{graphicx}
\usepackage[capitalise]{cleveref}
\usepackage{siunitx}

\usepackage{gensymb}
\usepackage{booktabs, etoolbox}
\usepackage{multirow}
\usepackage{subcaption}
\usepackage{floatrow}

\usepackage{scalerel}

\usepackage[T1]{fontenc}
\usepackage[font=small,labelfont=bf,tableposition=top]{caption}
\DeclareCaptionLabelFormat{andtable}{#1~#2  \&  \tablename~\thetable}

\newfloatcommand{capbtabbox}{table}[][\FBwidth]
\DeclareMathOperator*{\argmin}{argmin}   %
\DeclareMathOperator*{\argmax}{argmax}   %

\title{You Only Look at One: \\Category-Level Object Representations \\for Pose Estimation From a Single Example}

\author{
  Walter Goodwin, Ioannis Havoutis, Ingmar Posner\\
  Oxford Robotics Institute\\
  University of Oxford\\
  \texttt{firstname@robots.ox.ac.uk} \\
}

\begin{document}
\maketitle

\begin{abstract}
In order to meaningfully interact with the world, robot manipulators must be able to interpret objects they encounter. A critical aspect of this interpretation is pose estimation: inferring quantities that describe the position and orientation of an object in 3D space. Most existing approaches to pose estimation make limiting assumptions, often working only for specific, known object instances, or at best generalising to an object category using large pose-labelled datasets. In this work, we present a method for achieving category-level pose estimation by inspection of just a single object from a desired category. We show that we can subsequently perform accurate pose estimation for unseen objects from an inspected category, and considerably outperform prior work by exploiting multi-view correspondences. We demonstrate that our method runs in real-time, enabling a robot manipulator equipped with an RGBD sensor to perform online 6D pose estimation for novel objects.
Finally, we showcase our method in a continual learning setting, with a robot able to determine whether objects belong to known categories, and if not, use active perception to produce a one-shot category representation for subsequent pose estimation. 
\end{abstract}

\section{Introduction} \label{sec:intro}
\begin{figure}[h]
\centering
  \includegraphics[width=0.98\columnwidth]{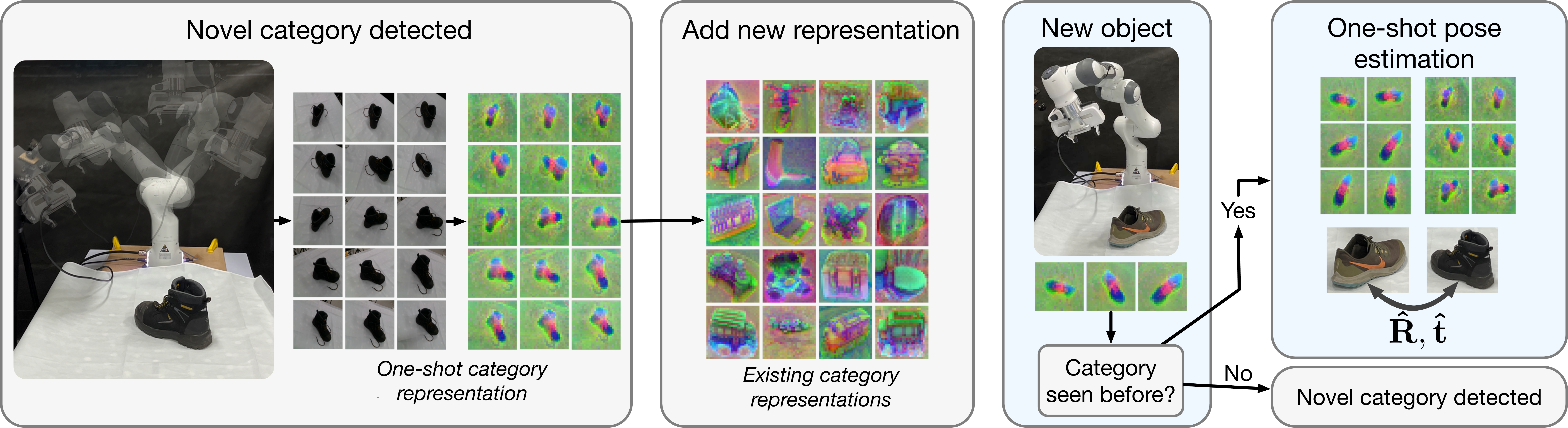}
  \caption{An overview of our method. Multiple views of an object from a new categories are taken and turned into a category-level representation in a ~\SI{10}{s} process. When a novel object is encountered, it is compared to all current representations. If it is deemed to belong to a previously captured category, our method performs 6D pose estimation in real-time. If it is not, it is scanned and a new category-level representation acquired. The approach runs in real-time and representations allow accurate pose estimation of unseen objects from the captured categories.}
  \label{fig:main}
\end{figure}

Many practical applications of robotics require that a robot is able to estimate the poses of objects that are encountered, that is, the parameters that describe their position and orientation in 3D space relative to some coordinate frame.  This capability is central to object manipulation, where to successfully rearrange objects, a robot needs to estimate object pose to infer how it differs from some goal state \cite{Wada2022Reorient}. Object pose estimation can also be helpful for mobile robots seeking to localise themselves in the world, by providing semantically grounded anchors \cite{Sucar2020, Ma2022}. 

Despite its central importance to robotics, approaches for object pose estimation remain limited in their abilities to generalise to novel objects. State of the art approaches often demand large labelled datasets or accurate CAD models just to produce pose estimators for single object instances \cite{Wang2019DenseFusion, He2020}. Over the last few years, attempts have been made to alleviate these issues by learning \textit{category level} pose estimators that train over multiple object \textit{instances} from a category and can generalise to unseen instances from that category \cite{Wang2019NOCS, Wang6Pack2020, Lin2021Pose}. Such generalisation to unseen objects is a much sought-after property in robotics, with some key examples being generalised manipulation \cite{Florence2018, Manuelli2019, Simeonov2021} and object-based SLAM \cite{Sucar2020}. However, these category-level approaches to pose estimation still have substantial data requirements in order to generalise - for instance, the Objectron dataset \cite{AhmadyanObjectron2021, Lin2021Pose} has over 17,000 object instances and 4 million frames to cover 9 object categories. 

Separately, some recent works have achieved very promising results on pose estimation for \textit{single} object instances by avoiding object-specific deep models, instead collecting a set of reference views of an object before performing robust feature matching for view retrieval and pose estimation \cite{YangTeaser2021, WenBundleTrack2021, ShugurovOSOP2022}. These approaches remove the need for large curated datasets required by deep learning approaches, but are designed for single object instances and thus fail to exhibit desirable category-level generalisation. 

In this work, we present an approach to object pose estimation that has the desirable properties of both of these paradigms. By leveraging the pre-trained features of a deep vision transformer (ViT) network \cite{Caron2021}, we show that we are able to leverage multiple views of a reference object to achieve pose estimation on \textit{any novel object from the same category}. Extending recent work on leveraging such features for pose estimation \cite{Goodwin2022}, we demonstrate much improved performance by also making use of multiple views of the novel target object. Further, in contrast to prior work, both the processing of object reference views into a reference object `model' suitable for downstream pose estimation, and the pose estimation process itself, can run in real-time and are thus suitable for in-the-wild robotic deployment. We demonstrate that our method drastically outperforms alternative methods in this one-shot category-level pose estimation setting, and show that this approach enables a robot manipulator to perform generalised manipulation tasks, in which a single `goal' pose for an object category is inspected by a robot manipulator, which then performs pick-and-place rearrangement to satisfy this goal pose when confronted with novel items from the same category. 

In summary, our main contributions are:
\begin{itemize}
    \item A training-free, multi-view approach to object pose estimation that only requires views of a single reference object, yet generalises to novel objects from the same category.
    \item A fast implementation of our pose estimation method that runs orders of magnitude faster than the closest baseline, enabling both real-time capturing of objects from novel categories, and real-time pose estimation for objects from known categories.
    \item Validation of the method's utility in a continual learning setting, in which the method is deployed tabula rasa and distinguishes novel categories from known categories, learning representations where appropriate.
\end{itemize}
\section{Related Work} \label{sec:related_work}

\subsection{Category-level object pose estimation}
Approaches to category-level object pose estimation can generally be divided into those which learn a category-level representation against which novel objects and images can be compared \cite{Wang2019NOCS, Chen2020, ChenSGPA2021, Wang2021_NeMo, Tian2020, ChenCASS2020}, those which leverage (a set of) CAD models for a category \cite{Shi2021, Xiao2019, Sahin2019, Grabner2018}, and those which train deep models to directly regress to parameters describing object pose \cite{GuptaPose2015, Xiao2021, ChenFSNet2021}. These methods require large pose-labelled datasets containing multiple instances of objects from each category of interest in order to generalise successfully. 

In contrast, in this work we build on a recent paper that proposes a novel zero-shot category-level pose estimation setting \cite{Goodwin2022}, in which, given a single image of some reference object and no pose-labelled data, the task is to estimate the pose of novel objects from the same category, using a handful RGB-D views. Building on \cite{Goodwin2022}, our work enables the use of multiple reference object views, and we show that this leads to a considerable improvement in performance. Further, while \cite{Goodwin2022} takes over 5 seconds to estimate pose for a single frame, our method can be run at over \SI{15}{Hz}, enabling real-time pose estimation on a robot manipulator.

\subsection{Template-based object pose estimation}
In making use of a collection of reference views of an object for pose estimation, our work is related to the literature on template matching. Template matching for pose estimation entails collecting many views of a reference object, and retrieving the closest view at run-time through a visual similarity measure. Recent work achieves impressive performance for single instances by constructing a point cloud of an object, and leveraging learnt 2D-to-3D correspondence matching and Perspective-n-Point (PnP) to recover pose when the object is next encountered \cite{SunOnepose2022}. This method just requires an object point cloud to perform pose estimation for novel objects. Similarly, \cite{ShugurovOSOP2022} uses a two-step matching process with dense CNN features, before a final PnP step is used to estimate object pose. Similar viewpoint-matching approaches have been used for pose tracking during manipulation \cite{WenBundleTrack2021}. \cite{LuContinual2022} solves a continual learning case, in which 3D object representations are captured when an object is first encountered by a robot, and subsequently matched against in a render-and-compare manner to provide 6D pose estimation. Template matching approaches have recently been employed in a few-shot setting, where following a pre-training stage on a diverse dataset, just a few views of a novel object are sufficient to estimate accurate pose \cite{HeFS6D2022}. All of the aforementioned methods enable pose estimation of novel objects provided that a collection of reference views, or a 3D model, can be acquired. However, the matching processes they employ are instance-specific, meaning that object templates cannot be used for pose estimation of previously unseen objects from related categories.

\subsection{Category-level object representations for manipulation}
Several prior works seek to learn category-level object descriptors to enable generalising robot manipulation skills from single object instances to broader object categories. \cite{Florence2018} learn dense object descriptors that can be trained in a mostly self-supervised manner from multiple views, and show that for several categories, after training on ~15 object instances nearest-neighbours in descriptor space between different objects tend to correlate with meaningful correspondences. Other work has sought to learn category-level \textit{keypoints} \cite{Manuelli2019}. Manipulation goals described as costs on keypoint positions can then be achieved on novel objects from known categories. More recently, category-level descriptors have been learnt on 3D object representations \cite{Simeonov2021}, by self-supervised training over a large set of CAD models from a category. These descriptors enable finding oriented correspondences between semantically equivalent points on objects from a training category, and have been used to place novel objects in target poses with a robot manipulator. These prior works require time-consuming training, over datasets collected from a not inconsiderable diversity of objects from a category of interest, and thus do not scale well. In this work, we show that dense object descriptors that are similarly able to produce meaningful part-based correspondences between objects from a given category can be achieved through inspection of just a single object instance in real-time. As in prior work, these descriptors can be used to match object poses, but in a considerably more scalable manner, with minimal data requirements. 
\section{Methods} \label{sec:methods}
\subsection{One-shot pose estimation setting}
To formalise the setting for one-shot pose estimation addressed in this work, we adopt and extend the notation from \cite{Goodwin2022}. We consider an object from a category $c$ to be represented by $M$ views, to produce a reference image set, $I_{\mathcal{R}_{1:M}}$. Except in certain ablations, we assume that depth images are also available from these views,  $D_{\mathcal{R}_{1:M}}$. We assume that camera extrinsics are known, a realistic assumption in most robotics contexts (e.g. a manipulator with a wrist-mounted camera), and denote these as $ \mathbb{T}_{\mathcal{R}_{1:M}}$ for the reference object views. At pose estimation time, a target object is encountered, and we assume that $N>1$ views of this object are also captured, with RGB images $I_{\mathcal{T}_{1:N}}$, depth images $D_{\mathcal{T}_{1:N}}$, and camera extrinsics $\mathbb{T}_{\mathcal{R}_{1:N}}$. We seek to recover a rotation $\hat{\mathbf{R}}$ and translation $\hat{\mathbf{t}}$ that describes the pose of the target object with respect to the reference object.

The reference object in the one-shot pose setting serves both as a model against which target object correspondences for pose estimation are made, but also to establish a canonical frame against which target object pose is measured. Previous work describes a similar setting in which just a single reference view is used as entailing `zero-shot' pose estimation, because this reference view is a bare minimum in order to render the pose-estimation problem well-posed. In our setting, because multiple reference views of an object are aggregated to form a representation for a category, we choose to describe this as one-shot pose estimation. Estimating the pose of a novel object \textit{relative} to the reference object is sufficient to generalise manipulation behaviours demonstrated on the reference object. 

\subsection{Descriptors and correspondences} \label{sec:method-corresp}
\cite{Goodwin2022} uses features extracted from a vision transformer network (ViT) for category-level pose estimation, showing that they generalise well across instances within a category. In this work we use the same ViT, trained with DINO \cite{Caron2021}, an unsupervised contrastive method that uses a multi-scale cropping process during training which, intuitively, encourages the network to discover correspondences between small local crops of an image, and a larger global crop. Empirically, DINO's features can identify objects but also object \textit{parts}. Further, ViTs take a patch-based approach to image processing which retains relatively high spatial resolution throughout the network, in contrast to CNNs, which spatially downsample such that later layers with expressive features lack spatial resolution. By using a ViT with a patch size of 8x8 pixels on RGB images resized to 224x224, the resulting feature maps at all layers are 28x28. This enables patch-based correspondences to be localised accurately in the image, and thus in 3D space through backprojection. We denote normalised feature maps extracted from image $I_{i}$ as $\Phi\left(I_{i}\right) \in \mathbb{R}^{28 \times 28 \times D}$, where D is the descriptor dimensionality (see \cref{sec:method-pca}). 

We construct correspondences between the reference and target objects by finding $P$ patch correspondences between every reference-target feature map pair $\{\Phi(I_{\mathcal{R}_{i}}), \Phi(I_{\mathcal{T}_{j}})\}_{i \in 1:M}^{j \in 1:N}$. For an image pair $I_{\mathcal{R}_{i}}, I_{\mathcal{T}_{j}}$ we denote these $\{u_{i_{k}}, v_{j_{k}}\}_{k\in1:P}$, where $u$, $v$ are indexes into the respective feature maps. The problem of finding strong correspondences between two feature maps $\Phi_1$, $\Phi_2$ is much studied, with many proposed solutions. Naive nearest neighbours can result in many-to-one matches that are physically implausible and thus not appropriate for downstream pose estimation. Mutual nearest neighbours \cite{Aberman2018} is a stronger condition, in which points $u$, $v$ are considered correspondences only if the closest descriptor to a point $u$ in $\Phi_1$ is $v$ (the standard nearest neighbours condition) \textit{and} the closest descriptor to $v$ in $\Phi_2$ is $u$. Formally, for $v=\operatorname{argmax}_{w} d(\Phi_{1_{u}}, \Phi_{2_{w}})$, we have $u'=\operatorname{argmax}_{w} d(\Phi_{2_{v}}, \Phi_{1_{w}})=u$, where $d(\cdot, \cdot)$ is the cosine similarity. Requiring this perfect `cycle' between the descriptors reduces spurious correspondences, but cannot guarantee that a certain number - or indeed any - correspondences will be found, which affects downstream pose estimation in our method. 

To ensure that $P$ correspondences are found for each reference-target pair, we use the cyclical distance metric proposed in \cite{Goodwin2022}, which can be thought of as a relaxation of mutual nearest neighbours matching: if there are $<P$ mutual nearest neighbours, additional correspondences are given points for which $u'-u$ as defined above is minimal. This acts as a spatial prior: correspondences that are \textit{almost} spatially consistent under the cycle are preferred: $u'$ and $u$ are likely to still belong to the same semantic \textit{part} of an object. We also experiment with a novel variant of this approach, in which the `cyclical distance' is measured in 3D world coordinates rather than pixel space, using the back-projected pixel coordinates (details in supplementary).

\subsection{Dimensionality reduction} \label{sec:method-pca}
In well-established approaches to learning category-level dense descriptors for robotic manipulation (e.g. \cite{Florence2018}), descriptor dimensionality tends to be low. The authors of \cite{Florence2018} note that for single instances, a 3-dimensional descriptor space tends to be sufficient to discriminate between object parts, though this requirement scales somewhat for multi-object descriptor networks. The raw ViT features we use as descriptors in this work are 384-dimensional. Such dimensionality is critical to the network's ability to satisfy the contrastive objective over the diverse dataset it is trained on. At the pose estimation stage, we would like to retain those factors of descriptor variation that enable us to distinguish different object parts and viewpoints, but it is natural to wonder whether there may be considerable redundancy in this descriptor space. To explore this, we perform PCA on the descriptors taken over all reference object views. The majority of descriptor variance is captured in a relatively small number of descriptor dimensions (supplementary). Further, we find that projecting features onto a small subset of principal components actually results in \textit{improved} pose estimation results. Certain results are show in \cref{tab:pca-dim} and discussed in \cref{sec:pca-res}. The category representations in \cref{fig:main} are the projections of descriptors onto the first three principal components.

In our experiments, we calculate principal components based on the \textit{reference} feature maps only, and use these to project both reference and target descriptors prior to finding correspondences. 

\subsection{Viewpoint estimation}
\subsubsection{Aggregating multi-view correspondences} \label{sec:mview-agg}
\begin{figure}[h]
\centering
  \includegraphics[width=\columnwidth]{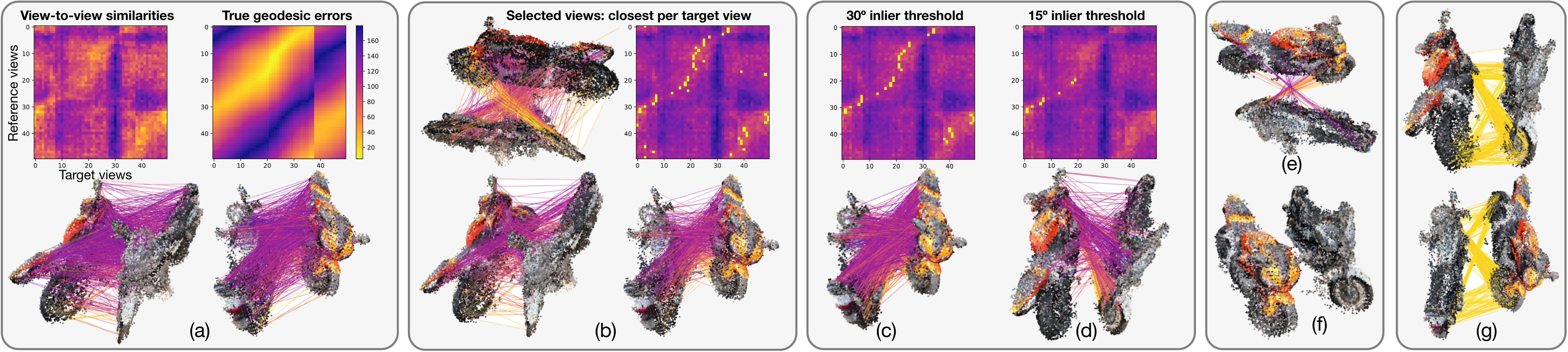}
  \caption{For visualisation, point clouds are created by a threshold on ViT attention masks to mask each view, aggregating, and removing outliers.
  \textbf{(a)} The matrices show (\textit{left}) the view-to-view similarities (sum of top-K correspondences) for each reference-target view pair (
  \protect\scalerel*{\includegraphics{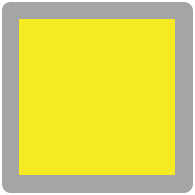}}{B}
  =high, 
  \protect\scalerel*{\includegraphics{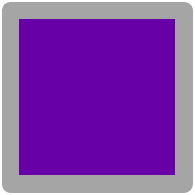}}{B}
  =low), (\textit{right}) the true geodesic error between the object poses in these views, which can be seen to be correlate with the similarities. Below, a random subset of correspondences across all view pairs is visualised. 
  \textbf{(b)} For each target view, the corresponding reference view with the highest similarity is chosen, with these pairs highlighted in yellow on the matrix. A subset of the resulting filtered correspondences is shown.
  \textbf{(c)} The resulting inlier frames by consensus filtering with a threshold of 30\degree (\cref{sec:mview-agg}), and correspondences.
  \textbf{(d)} The same with a tighter inlier threshold of 15\degree.
  \textbf{(e)} The inlier correspondence set following RANSAC.
  \textbf{(f)} Objects aligned by estimated pose, offset by a translation for clarity.
  \textbf{(g)} Taking top-K correspondences globally from the whole matching process (rather than filtering to good views) leads to conflicting and erroneous correspondences.}
  \label{fig:correspond}
\end{figure}
Following \cite{Goodwin2022}, we assign similarities $\mathcal{S}_{ij}$ to reference-target view pairs e.g. $I_{\mathcal{R}_{i}}, I_{\mathcal{T}_{j}}$  as the sum of the similarities between the top-K correspondences for this pair, $\{u_{i_{k}}, v_{j_{k}}\}_{k\in1:P}$, found as described previously. That is, $\mathcal{S}_{ij} = \sum_{k = 1}^{P} d(u_{i_{k}}, v_{j_{k}})$. Considering all reference images $I_{\mathcal{R}_{i=1:M}}$ and target images $I_{\mathcal{T}_{j=1:N}}$, we arrive at a view similarity matrix $\mathbb{S}_{M\times N}$, with elements $\mathbb{S}_{M\times N}[i,j]=\mathcal{S}_{ij}$. An example of such a similarity matrix, and subsequent processing steps and their effect on the correspondence set, is shown in \cref{fig:correspond}. We leverage this matrix to filter to a subset of best-fit view pairs. First, for each target frame we retain just the reference frame with maximum similarity. Subsequently, we further filter based on the estimated relative orientation estimates we would arrive at if we took each pair to represent a \textit{perfect} alignment. With this set of relative poses, we find the largest consensus set by sampling many random poses in SO(3) and finding the viewpoint pairs with a relative pose within some threshold $\theta$\degree distance from this pose. These largest such set is chosen.
Correspondences from these remaining views are used in the final rigid body transform solution.

\subsubsection{Final pose estimation} \label{sec:umeyama}
Given a filtered subset of reference-target view pairs which have a consensus on an approximate SO(3) pose prediction, we seek to estimate a refined SO(3) pose, along with a translation, to produce a full 6D object pose estimate, and a scaling, to handle intra-category size differences. For this, we assume that the reference and target objects are related by such a 7-D rigid-body transform, and solve for this using RANSAC and Umeyama's method \cite{Umeyama1991}. With $K$ correspondences per view pair and a subset of $Q$ view pairs, we have correspondences $\{u_{k},v_{k}\}_{k=1:QK}$. Using reference and target depth maps $D_{\mathcal{R}_{i=1:M}}$ and $I_{\mathcal{T}_{j=1:N}}$, we back project these pixel coordinates to 3D points $\{\mathbf{u}_{k},\mathbf{v}_{k}\}_{k=1:QK}$. For each RANSAC trial we sample four pairs of corresponding points as required by Umeyama's method, and the estimated rotation $\hat{\mathbf{R}}$, translation $\hat{\mathbf{t}}$ and scaling $\hat{\lambda}$ satisfy 
\begin{equation}
    (\hat{\lambda}, \hat{\mathbf{R}}, \hat{\mathbf{t}}) = \argmin_{(\lambda, \mathbf{R}, \mathbf{t})} \sum_{k = 1}^{4} \mathbf{v}_{k} - (\lambda \mathbf{R}\mathbf{u}_{k}+\mathbf{t})
\end{equation}
We run RANSAC for 1000 trials and make a final estimate using the largest resulting inlier set.

\subsection{Continual category learning} \label{sec:method-ncd}
An often overlooked upstream requirement of most object pose estimation methods used in robotics is object detection and classification. Many approaches to category-level \textit{and} instance-level pose estimation require knowledge of the target object's category or identity to choose the appropriate network \cite{Lin2021Pose}, network head \cite{Xiang2018}, or template \cite{SunOnepose2022}. We would like to be able to leverage the one-shot setting presented in this work in a fully autonomous context, where the category identities of objects encountered by a robot are estimated for subsequent pose estimation. Specifically, a compelling setting in robotics is that of continual learning, where a robot encountering novel objects can either assign them to categories seen before, or determine that they are a novel category. In our case, the former would lead to a selection of a suitable reference object for pose estimation, while in the latter, the robot would employ active perception to produce multiple views of this novel object, forming a representation for a novel category. We denote the set of previously `discovered' categories $\mathcal{C}_{1:S}$, and given one or more views of a novel object $I_{\mathcal{T}_{1:N}}$, wish to determine the object's identity $\hat{\mathcal{C}} \in \{C_1...C_{S+1}\}$, where $C_{S+1}$ would imply this object belongs to an unseen category. We seek a similarity metric that enables fast checking against a potentially large number of reference objects. For this, we use the [CLS] tokens from the last layer of the DINO ViT \cite{Caron2021}. For an existing category, we represent it as the set of [CLS] tokens over all sequences and views that have been assigned to it, such that the whole feature set is $\{\Psi_{s}={\psi_{s_{1}},...,\psi_{s_{F}}}\}_{s=1:S}$. For a new category with [CLS] tokens over its views $\psi_{x_{1}},...,\psi_{x_{G}}$, we assign the category $s$ as:
\begin{equation}
    \argmax_{s} f(\Psi_{s}, \Psi_{x}) \text{ if } f(\Psi_{s}, \Psi_{x}) > \theta\text{, where }
    f(\Psi_{s}, \Psi_{x})=\big(\sum_{i=1}^{F}\sum_{j=1}^{G}d(\phi_{{s}_j}, \phi_{{x}_i})\big)
\end{equation}
Where $\theta$ is a threshold on similarity (details in supplementary). If no category meets this threshold, the object is assumed to belong to a novel category.

\section{Experimental Results} \label{sec:exps}
\begin{table}[!htb]
    \small
    \addtolength{\tabcolsep}{2pt}
    \centering
    \caption{Pose estimation accuracy (orientation). We report Acc@$\theta^{\circ}$, with $\theta\in\{30^{\circ},15^{\circ},7.5^{\circ}\}$, giving the percentage of estimates that fall a certain maximum threshold on geodesic error, and the median error `Med. Err' (per category, then averaged). We compare to TEASER++ \cite{YangTeaser2021}, and Goodwin2022 \cite{Goodwin2022}. Suffixes on our method ablations: \textbf{R}: retrieval; pose estimated based on mean of most similar views. \textbf{C}: consensus, iterative removing far-from-mean views. \textbf{U}: Umeyama's method; rigid body solution using best-view correspondences.  \textbf{D}: descriptor dimensionality reduction (to 32 components). Methods use 10 reference and target views (Goodwin2022 uses just 1 reference). \textbf{+} indicates 30 views used.}\label{tab:main_res}
    \resizebox{\linewidth}{!}{ %
    \begin{tabular}{ccccccccccc}
    \toprule
    \multicolumn{1}{l}{} & \multicolumn{4}{c}{All Categories} & \multicolumn{6}{c}{Per Category (Acc@30º), \%}\\
    \cmidrule(rl){2-5}
    \cmidrule(rl){6-11}
    Method & Med. Err ($\downarrow$) & Acc30º ($\uparrow$) & Acc15º & Acc7.5º & B'pack & Car & Chair & Keyboard & Laptop & M'cycle \\
    \midrule
    TEASER++ \cite{YangTeaser2021} & 126.4 & 3.75 & 1.1 & 0.3 & 1 & 5 & 9 & 8 & 6 & 1 \\
    Goodwin2022 \cite{Goodwin2022} & 47.7 & 49.4 & 28.35 & 9.6 & 44 & 65 & 47 & 69 & 85 & 85 \\
    \midrule
    Ours-R    & 58.0 & 29.9 & 13.4 & 4.4 & 44 & 23 & 34 & 15 & 63 & 73 \\
    Ours-RC   & 32.7 & 56.6 & 35.0 & 13.1 & 64 & 74 & 58 & 57 & 91 & 82 \\
    Ours-U    & 36.9 & 59.4 & 45.7 & 26.6 & 49 & \textbf{92} & 59 & \textbf{80} & 97 & \textbf{100} \\
    Ours-UC   & 31.9 & 61.5 & 47.5 & 27.4 & 61 & 78 & 64 & 67 & 98 & 99 \\
    Ours-UCD  & 26.2 & 64.5 & 49.4 & 29.2 & 60 & 87 & 70 & 70 & 99 & 99 \\
    Ours-UCD+ & \textbf{21.4} & \textbf{69.8} & \textbf{54.6} & \textbf{34.5} & \textbf{67} & 86 & \textbf{76} & 76 & \textbf{100} & \textbf{100} \\
    \bottomrule
    \end{tabular}
    }

\end{table}

\subsection{Multi-view one-shot pose estimation} \label{sec:exps_multi}
To evaluate the performance of the proposed pose estimation method on a diverse range of categories and instances, we use a pose-labelled subset \cite{Goodwin2022} of the Common Objects in 3D (CO3D) dataset \cite{ReizensteinCO3D2021}, for which the ground truth relative pose between video sequences of 10 distinct objects from each of 20 categories is labelled. Each sequence contains approximately 100 frames, and is taken by a smartphone in a hand-held turntable-style object scan. Camera extrinsics and depth images in this dataset are approximate, being recovered from RGB video sequences by a Structure-from-Motion approach \cite{Schoenberger2016}. While we find that (with appropriate real-time depth completion, described in supplementary) the extrinsics and depth are of sufficiently high quality, we consider this dataset to be a lower bound on data quality from a wrist-mounted RGB-D camera. 

The results of our method and certain ablations are shown in \cref{tab:main_res}. We compare to the method of \cite{Goodwin2022}, which uses a single reference image, and to TEASER++, a fast point cloud registration algorithm that works in the presence of a large number of outliers, unknown correspondences, and - unlike ICP - does not require a strong initial guess. Further, it leverages an explicit bound on the noise it assumes present in the \textit{non-outlier} points, which is analogous to the inlier threshold we use in our RANSAC stage (see supplementary), and thus forms an interesting baseline for leveraging \textit{purely} geometric correspondences for category-level pose estimation.

The improvement over \cite{Goodwin2022} can be seen especially strongly in at the more precise Acc@15º and Acc@7.5º, where our full method demonstrates almost double and triple the accuracy respectively, and scores 20.4\% higher on Acc@30º. This underlines the importance of aggregating many-to-many view comparisons to arrive at pose estimates. While \cite{YangTeaser2021} shows better-than-random performance for most categories, it is not in general able to solve the category-level pose estimation problem, indicating that stronger priors than just geometric distances are necessary in this setting. Further results, including those for translation errors, are in the supplementary.

\subsubsection{Effect of descriptor dimensionality reduction} \label{sec:pca-res}
In \cref{sec:method-pca}, we discuss our approach to reducing descriptor dimensionality. \cref{tab:pca-dim} shows the impact of using differing numbers of principal components on the performance of our pose estimation method. We find that projecting descriptors onto a small subset of principal components not only serves to reduce both computation time, and memory usage per reference object, but also have a beneficial effect on performance. For 10, 20 and 30 view regimes, we see 8\%, 2\% and 3\% improvements respectively when reducing from 384D to 32D descriptors. An intuitive explanation for the power of retaining just the top principal components from multi-view features is that doing so discards directions in feature space which are \textit{not} variant under viewpoint. For instance, a ViT feature for a patch containing a wing-mirror of a red motorbike might capture the semantic nature of this part - which is important spatially for predicting pose - but might also capture the redness: the latter is invariant over the views, so would not feature in the first principal components. This would be an example of an instance-specific property that might be encoded in the full feature dimensionality, yet would be detrimental to finding strong cross-instance correspondences.
\begin{table}[!htb]
\small
\addtolength{\tabcolsep}{2pt}
\centering
\caption{Pose estimation (orientation) performance over 20 object categories, with varying descriptor dimensionality reduction and number of reference and target views. Dimensionality reduction by an order of magnitude (from 384D to 32D or 24D) both reduces computational burden \textit{and} improves results.}%
\label{tab:pca-dim}
\begin{tabular}
{cccccccc}
\toprule
Views & \# Dim & Med. Err & Acc@30º & Acc@15º & Acc@7.5º & Time (ms) & Memory (MB) \\
\cmidrule(rl){1-2}
\cmidrule(rl){3-6}
\cmidrule(rl){7-8}
10 & 384 & 25.8 & 61.4 & 39.7 & 17.2 & 2.96 & 12.0 \\
10 & 32 & 23.7 & 66.2 & 43.0 & 18.3 & 2.60 & 1.0 \\
10 & 6 & 32.0 & 51.6 & 29.5 & 11.2 & 2.60 & 0.2 \\
10 & 3 & 59.2 & 35.8 & 17.1 & 5.9  & 2.60 & 0.1 \\
20 & 384 & 22.1 & 68.4 & 49.6 & 25.9 & 4.06 & 24.1 \\
20 & 32 & 22.0 & 69.8 & 50.1 & 24.0 & 3.36 & 2.0 \\
30 & 384 & 20.4 & 69.9 & 51.4 & 25.4 & 5.18 & 36.1 \\
30 & 32 & \textbf{20.3} & \textbf{72.1} & \textbf{52.4} & \textbf{26.2} & 4.23 & 3.0 \\
\bottomrule
\end{tabular}
\end{table}

\vspace{-2mm}
\subsection{Novel category discovery} 
Having verified that our method can produce accurate pose estimation for a range of objects using just a single reference object for each category, we note that in the setting above, we assume that the object identities are known. That is, we never attempt to predict a bicycle's pose from a teddy-bear reference. Our approach for continual learning of categories (\cref{sec:method-ncd}) should enable such identity matching (classification and novel category discovery) to be fully autonomous. To assess this, we simulate 1,000 continual learning episodes using 10 sequences from each of 20 CO3D categories. In each episode, 250 sequences are drawn sequentially and at random. For each sequence we either assign it to an existing or a novel category as described in \cref{sec:method-ncd}. Objects that are erroneously believed to be novel categories are False Negatives (FN), and those assigned to a set where they are not the majority true category are False Positives (FP). Through tuning the similarity threshold, a 0\% FP rate can be reached with a 26\% FN rate. 
\vspace{-1.5mm}
\subsection{Robotic novel object pose rearrangement} \label{sec:exps_robot}
\vspace{-1.5mm}
We deploy our method on RealSense D435i camera wrist-mounted on a Panda robot. To scan objects from novel categories, we sample 10 or 20 views in an approximate hemisphere above the object, with the camera oriented towards the object. When an object is detected as belonging to a known class (\cref{sec:method-ncd}), the robot traverses between several random waypoints that look at the object, and we inspect pose estimates through a visualised oriented 3D bounding box. Further details are in the supplementary.

\vspace{-2mm}
\section{Limitations} \label{sec:limits}
\vspace{-2mm}
Our work assumes that camera extrinsics are available between frames at runtime, and for best performance also assumes that a depth camera is available. While these are both readily satisfied in the motivating example of a robot manipulator with a wrist-mounted camera, these conditions may not be so readily available in, for example, an AR context in which a moving mobile phone is the candidate device for running pose estimation. However, the correspondence component of our method can still be used in a depth-free setting (see \cref{sec:exps_multi}), with either Perspective-n-Point if depth is still available for the \textit{reference} object, or essential matrix estimation if depth is available for neither. 
This work assumes that the relationship between different objects within a category can be captured reasonably well by a 7-parameter rigid-body transform between the spatial locations of corresponding salient points on the objects. This would not hold for deformable or articulated objects, or objects categories exhibiting drastic shape diversity. Relaxing this assumption would be a promising direction for future work. An extension to the proposed method which models inter-instance relationships with a 9-D affine transform is described in the supplementary.

\vspace{-1.5mm}
\section{Conclusion} \label{sec:conclusion}
\vspace{-2mm}
In this work we propose an approach to object pose estimation that runs in real-time on RGB-D data, and can provide accurate pose estimates at the category level by leveraging multiple views of a single reference object from the category. We show that high dimensional features from a pre-trained ViT can become high quality and efficient descriptors through dimensionality reduction. We demonstrate that our method can function as a fully autonomous approach to continual learning of category representations for pose estimation, with a scalable approach to classification of novel objects. Our method runs at over \SI{15}{Hz} on a camera wrist-mounted on a robot arm. 

\acknowledgments{The authors gratefully acknowledge the use of the University of Oxford Advanced Research Computing (ARC) facility \url{http://dx.doi.org/10.5281/zenodo.22558}. We thank Jack Collins, Oiwi Parker Jones, and Sagar Vaze for looking at early drafts.}

\bibliography{multipose.bib}  

\clearpage
\maketitle
\section{Appendix}
\renewcommand{\thefigure}{A.\arabic{figure}}
\setcounter{figure}{0} 
\renewcommand{\thetable}{A.\arabic{table}}
\setcounter{table}{0} 

In this supplementary material, we cover certain aspects of our implementation in more detail. We present plots visualising the first descriptors extracted for various object categories. We describe the extension of our method to fitting 9-D transforms (rather than the 7-D similarity transform) between object instances. Finally, we give details of our experiments on a Panda robot arm with a RealSense camera, which demonstrated that the method could run at 15fps while providing pose estimates for previously unseen objects.

\subsection{Implementation details}

\subsubsection{Descriptor dimensionality reduction}
\begin{figure}[htb!]
\centering
  \includegraphics[width=\columnwidth]{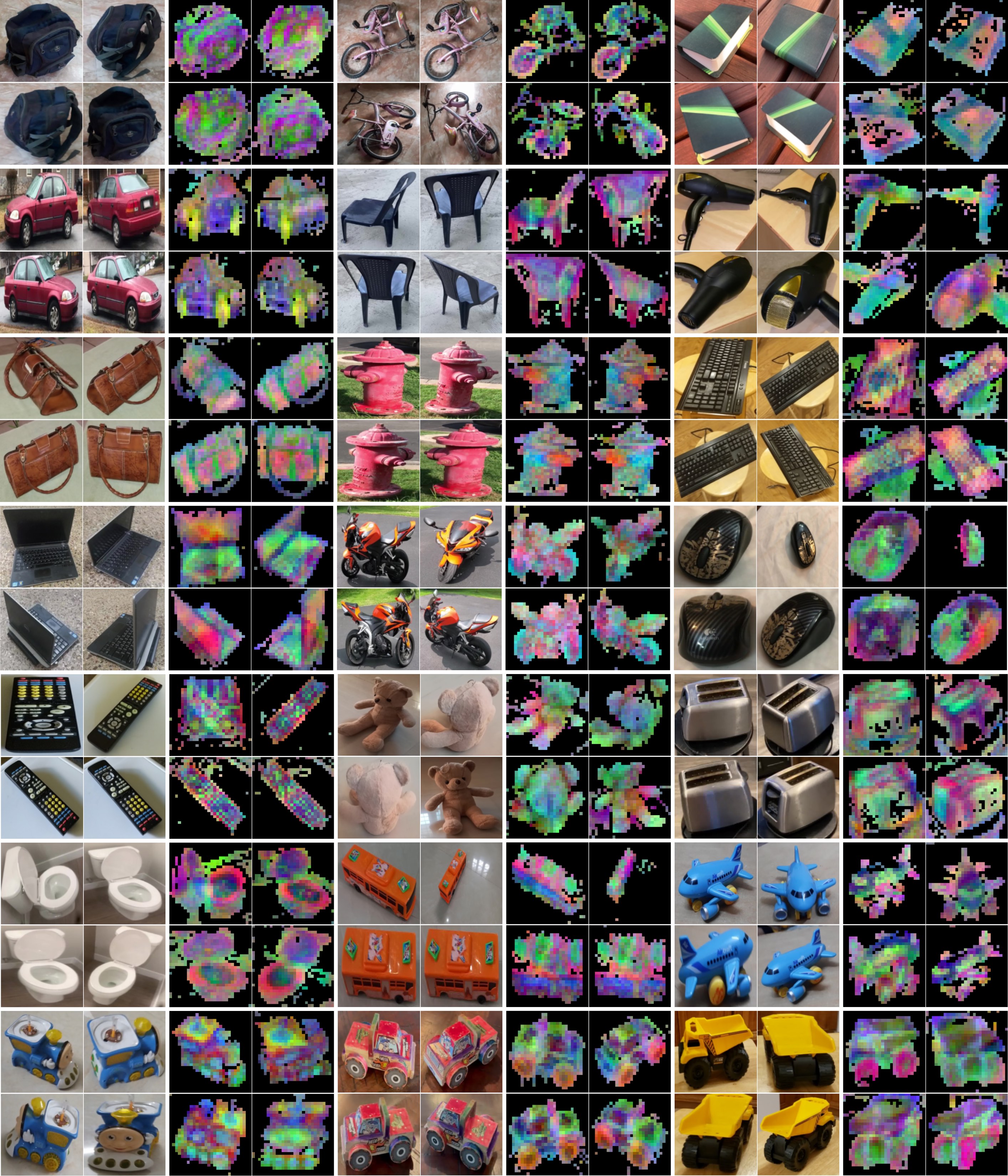}
  \caption{Examples from the CO3D dataset, showing the first three principal components for target sequences from each of 20 considered categories. For each example, the principal components have been calculated on a \textit{different} reference sequence. PCA feature maps are masked by a threshold on saliency computed from the ViT attention maps.}
  \label{fig:pca-all-cats}
\end{figure}

\begin{figure}[htb!]
\centering
  \includegraphics[width=\columnwidth]{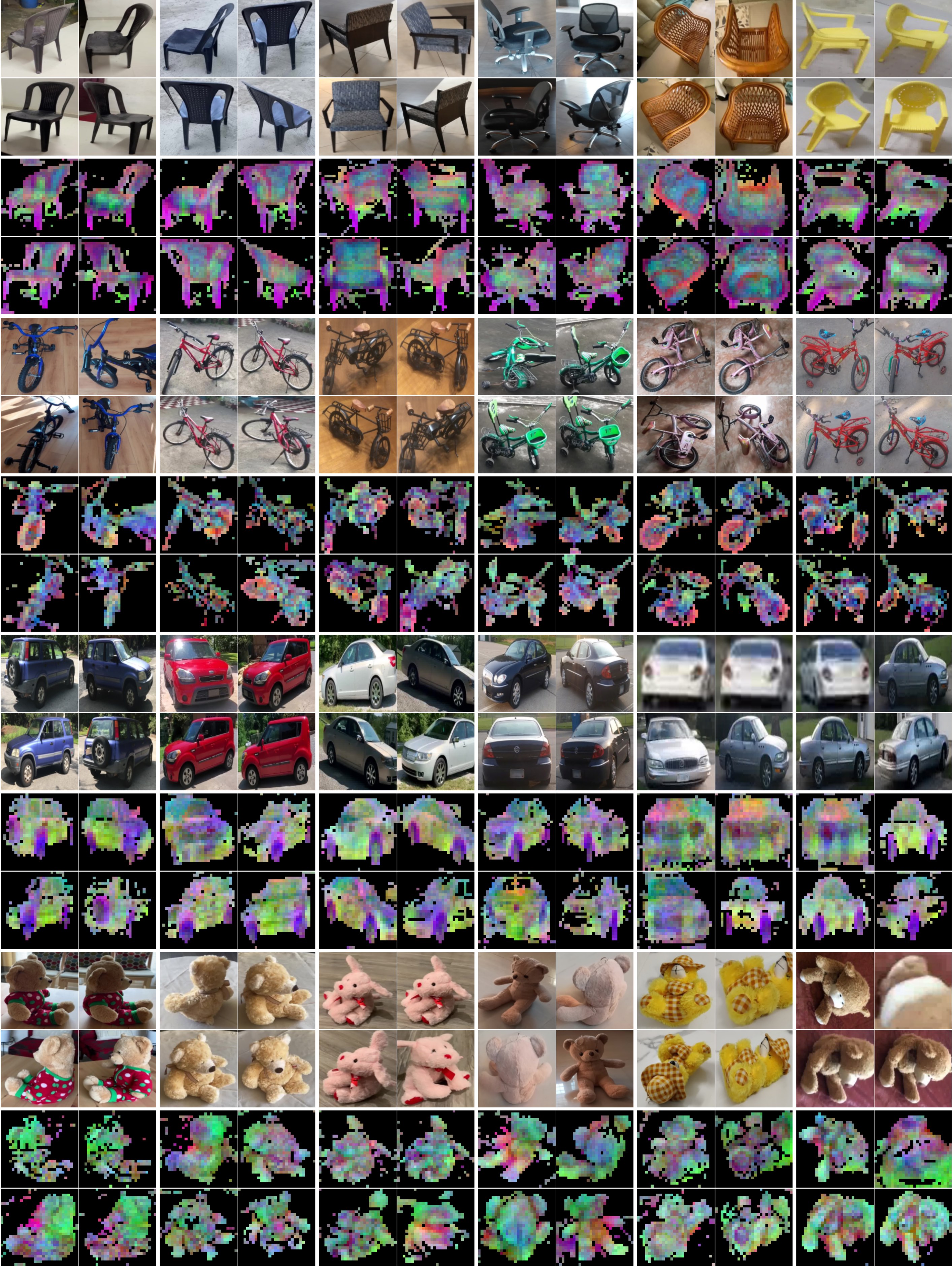}
  \caption{Visualising the first three principal components calculated from the features of a single reference sequence (leftmost images for each category) on five further sequences from the same category. Consistent colouring of corresponding parts can be seen. This property of invariance to varying instances within a category is exploited by our method for robust cross-instance correspondence estimation, used in our pose estimation method.} 
  \label{fig:pca-multi-seq}
\end{figure}

In \cref{sec:method-pca}, we describe our approach to using principal components analysis (PCA) to reduce descriptor dimensionality. In \cref{sec:pca-res}, we show that empirically, such dimensionality reduction actually improves performance up to a point (an over 10-fold decrease in dimensionality from 384-D raw ViT features to 32-D descriptors improves performance considerably for pose estimation over the CO3D dataset). In this appendix, we provide illustrative examples of the descriptors resulting from PCA. \cref{fig:pca-all-cats} shows the first three principal components for several frames from each of 20 CO3D categories. The components onto which these frames' descriptors have been projected have been, in each case, calculated on a distinct reference sequence from the same category. Descriptor plots are masked by a threshold on ViT saliency for clarity. \cref{fig:pca-multi-seq} also shows the projection of sequence descriptors onto the first three principal components, but shows the result of using principal components calculated on descriptors from a single reference sequence (the left-most sequence for each category) on five other distinct sequences drawn from the same category. The cross-instance generalisation of the DINO ViT features, and of descriptors derived from these, can be seen in the consistent colouring (in the space of the first three principal components) of key object parts across diverse instances.

\subsubsection{Fast depth completion}
In both our static datasets (\cref{sec:exps_multi}) and robot manipulator settings \cref{sec:exps_robot}, depth images are incomplete. For fast and scene-agnostic depth completion, we take inspiration from \cite{Ku2018}, which proposes a simple sequence of kernel-based filters for fast depth completion. We use a similar sequence of kernels to process our depth images. Of particular importance is that there are no holes in the depth image, as these can cause numerical instabilities when back-projected points are used to estimate rigid-body transforms in our method. As a final stage, we fill any remaining holes with a dilation operation with a very large kernel size, and set any remaining empty values to the mean depth over the image. Depth processing takes \SI{4.1}{\milli\second}. An example result on a sequence from the RealSense camera can be seen in \cref{fig:bbox_proc}.

\subsubsection{TEASER++ baseline}
For the TEASER++ \cite{YangTeaser2021} baseline experiments reported in \cref{sec:exps}, we use the official implementation from \url{https://github.com/MIT-SPARK/TEASER-plusplus} with all parameters at defaults. 

\subsection{Additional experiments}
\subsubsection{3D cyclical distance for correspondence matching}
In \cref{sec:method-corresp}, we describe the cyclical distance metric for selecting strong correspondences from descriptor similarity matrices, which is introduced in \cite{Goodwin2022}. This cyclical distance metric for selecting correspondences can be thought of as a spatial prior: correspondences that form a cycle from reference to target back to reference image patch (by nearest neighbours) which arrives at a \textit{close} location in patch-space to the original location are more likely to be meaningful, because a close patch location could very likely contain the same part of an object as the original patch. Patch distances, though, are a 2D measure, while in this work we have access to depth maps. We experiment with a 3D version of the cyclical distance measure in which correspondences are ranked based on the distance \textit{in actual 3D space} between the original patch and the final patch. We found a small improvement from this (+0.6\% on Acc30 across CO3D in a 10 reference, 10 target image setting), but we do not believe this result to be statistically significant.

\subsubsection{Replacing 7D similarity transform with 9D affine transform}
As noted in \cref{sec:limits}, the use of a 7D similarity transform (1D isotropic scale, and 6D pose) to model the relationship between correspondences found between two object instances from a category is almost certainly over-prescriptive. In this section, we how a 9D affine transform can be used instead with very few changes to the method. This is a more general model, and has promise to be more suitable for certain categories. In 9D setting, rather than a single isotropic scaling parameter $\lambda$, we seek a vector $\mathbf{s}\in\mathbb{R}^{3}$ which models separate scalings for each dimension.

Otherwise following \cref{sec:umeyama}, we seek a rotation $\hat{\mathbf{R}}$, translation $\hat{\mathbf{t}}$ and scaling $\hat{\mathbf{s}}$ that, for $K$ corresponding 3D points $\{\mathbf{u}_{k},\mathbf{v}_{k}\}_{k=1:K}$ satisfy the following: 
\begin{equation}
    (\mathbf{s}, \hat{\mathbf{R}}, \hat{\mathbf{t}}) = \argmin_{(\mathbf{s}, \mathbf{R}, \mathbf{t})} \sum_{k = 1}^{K} \mathbf{v}_{k} - (\mathrm{diag}(\mathbf{s})\mathbf{R}\mathbf{u}_{k}+\mathbf{t})
\end{equation}

In the 7D case ($\lambda$ scaling rather than $\mathbf{s}$), Umeyama's method gives a fast and closed form solution that scales well with the number of points as it is based on the singular value decomposition of the covariance matrix of the two correspondence matrices \cite{Umeyama1991}. 

A similar method has been proposed in \cite{Awange2008} to calculate the 9D transform described above. We refer the reader to that paper for further details. We implemented this method to evaluate whether finding a 9D transform might lead to better pose estimates by allowing for a more accurate category-level model of spatial correspondence. As with Umeyama's method, the most time-consuming component of this algorithm is computing the singular value decomposition of the covariance matrix, and we are still able to run 1,000 RANSAC trials in a few milliseconds with this method and CUDA acceleration. 

While we did not find this method to improve pose estimation performance on aggregate over the categories used in this work (a -0.5\% drop in Acc30 across CO3D in a 10 reference, 10 target image setting), this may be because the categories considered tend to have objects whose shape relationship is captured acceptable by a single scaling factor.

Finally, while this work's evaluation is on pose estimation, the underlying methodology and finding of robust category-level correspondences could be applied to the challenging setting of category-level grasping. In this context, there is good evidence that the ability to infer non-uniform scaling is important for transferring grasps between items within a category. Recent works have motivated the use of the 9D transform described here in representing a category-level canonical space for objects \cite{Wen2021, Wen2022}, extending Normalised Object Canonical Space (NOCS) \cite{Wang2019NOCS} to Non-Uniform NOCS (NUNOCS). 

\subsubsection{RANSAC inlier threshold}
We use RANSAC to find an optimal transform as described in \cref{sec:umeyama}. For the results in \cref{tab:main_res}, we used an inlier threshold of 0.2 for all categories, in order to match the conditions in \cite{Goodwin2022}. All objects in the dataset are at the same scale, measuring about 6 units long on their longest side. This threshold is thus intuitively quite restrictive: a point on the reference object under the estimated transform must not be more than about 3.5\% of the object's overall size away from the corresponding point on the target object, or else it will not be counted as an inlier. We experimented with various inlier thresholds, and found that setting this parameter to be higher, based on the above intuition, has a positive effect. The results in \cref{tab:pca-dim}, for instance, use a threshold of 0.5 (c.f. 72.1\% Acc@30\degree for the 30-vs-30 image results, vs 69.8\% Acc@30\degree in \cref{tab:main_res}).

\subsection{Further results}
\subsubsection{Translation estimation performance}
\cref{tab:trans_res} reports results for estimating the translation component of 6D pose.
\begin{table}[!htb]
    \small
    \addtolength{\tabcolsep}{2pt}
    \centering
    \caption{Pose estimation accuracy (translation). We report Acc@$\delta$, with $\delta\in\{1.0, 0.5, 0.2\}$, giving the percentage of estimates that fall within a certain maximum threshold on Euclidean distance from ground truth, and the median error `Med. Err' (per category, then averaged). These numbers are absolute, as every category in the CO3D dataset is at the same scale, with the longest side scaled to $2\pi$. Thus, a translation error of 1.0 is approximately 1/6 the longest side of the object. Suffixes on our method ablations: \textbf{C}: consensus by largest inlier group  \textbf{U}: Umeyama's method; rigid body solution using best-view correspondences.  \textbf{D}: descriptor dimensionality reduction (to 32 components). Methods use 10 reference and target views. \textbf{+} indicates 30 views used.}\label{tab:trans_res}
    \resizebox{\linewidth}{!}{ %
    \begin{tabular}{ccccccccccc}
    \toprule
    \multicolumn{1}{l}{} & \multicolumn{4}{c}{All Categories} & \multicolumn{6}{c}{Per Category (Acc@0.5), \%}\\
    \cmidrule(rl){2-5}
    \cmidrule(rl){6-11}
    Method & Med. Err ($\downarrow$) & Acc@1.0 ($\uparrow$) & Acc@0.5 & Acc@0.2 & B'pack & Car & Chair & Keyboard & Laptop & M'cycle \\
    \midrule
    Ours-U    & 0.586 & 76.1 & 46.9 & 19.9 & 25 & \textbf{97} & 27 & 80 & 96 & 85 \\
    Ours-UC   & 0.548 & 78.3 & 49.4 & 19.1 & 37 & 84 & 36 & 81 & 99 & 84 \\
    Ours-UCD  & 0.498 & 81.6 & 52.6 & 20.7 & \textbf{39} & 91 & 35 & \textbf{86} & 99 & 85 \\
    Ours-UCD+ & \textbf{0.489} & \textbf{81.8} & \textbf{53.9} & \textbf{22.3} & 36 & 90 & \textbf{40} & 82 & \textbf{100} & \textbf{90} \\
\bottomrule
    \end{tabular}
    }

\end{table}

\subsubsection{Visual examples of inlier correspondences following RANSAC}
\begin{figure}[htb!]
\centering
  \includegraphics[width=\columnwidth]{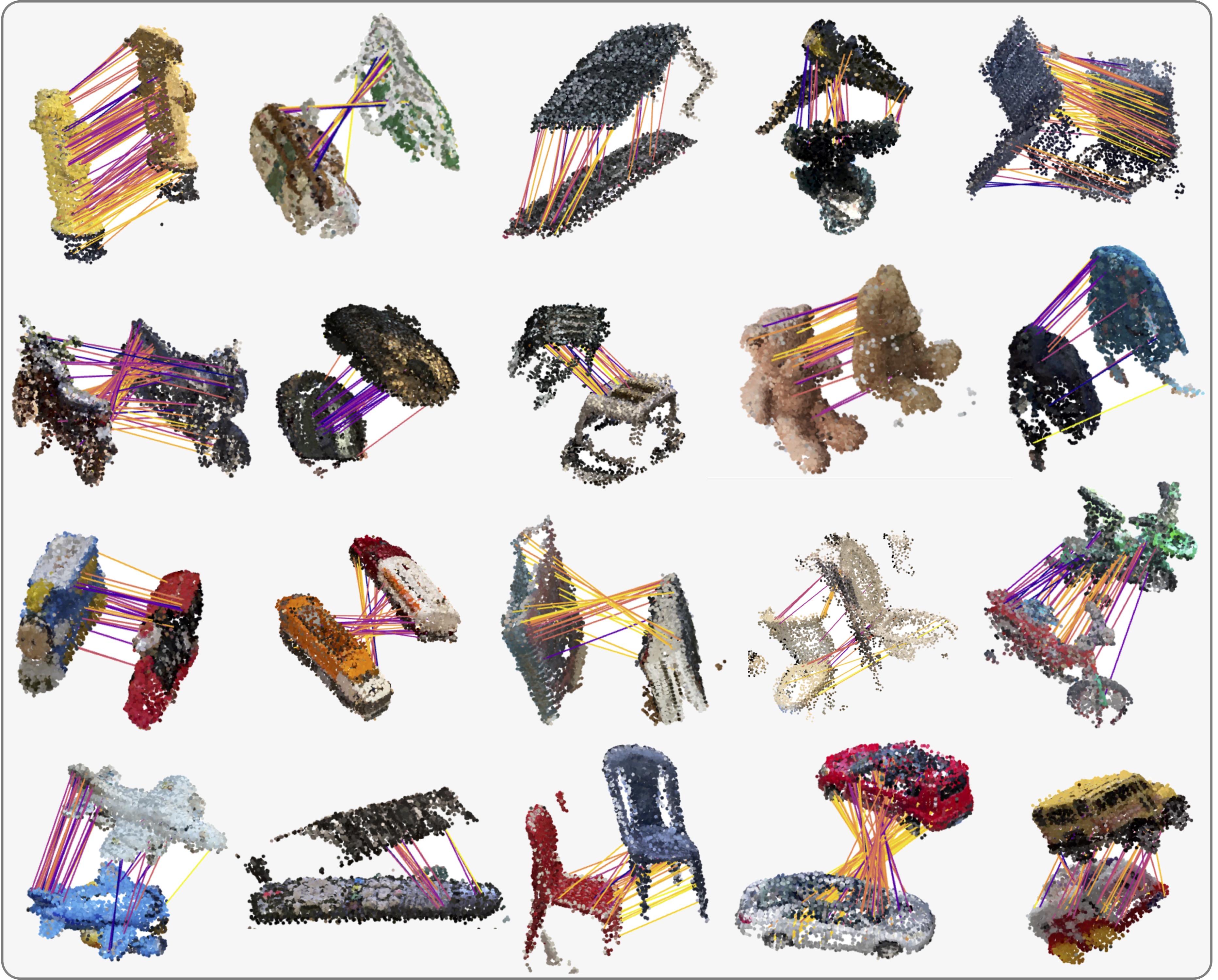}
  \caption{An example reference-target object pair from each of 20 CO3D categories used for pose estimation. Point clouds are created from the images and depth maps as described in \cref{sec:bbox}. Point clouds are rendered with respect to the camera viewpoint of the first frame in their respective sequences, with an added fix offset to avoid overlap. Lines between the objects show the \textit{inlier set} of correspondences following the RANSAC rigid body solution (\cref{sec:umeyama}). As in \cref{fig:correspond}, line colour shows correspondence similarity 
  (
  \protect\scalerel*{\includegraphics{figures/glyphs/yellow.pdf}}{B}
  =higher, 
  \protect\scalerel*{\includegraphics{figures/glyphs/purple.pdf}}{B}
  =lower). Categories, from left to right, top to bottom: \textit{hydrant, handbag, keyboard, hairdryer, laptop, motorbike, mouse, toaster, teddybear, backpack, toy train, toy bust, book, toilet, bicycle, toy plane, remote, chair, car, toy truck}.}
  \label{fig:inlier-plot}
\end{figure}

\cref{fig:inlier-plot} visualises a key part of the pose estimation process - the inlier set following RANSAC - on an example reference-target object pair from each of the 20 CO3D categories considered.

\subsection{Robot experiments}
In \cref{sec:exps_robot}, we describe the setting for deploying our pose estimation method on a Panda robot arm with a wrist-mounted RealSense camera. In this appendix, we expand on several fully automated pre-processing steps that are important implementation details for this real-world setting. For videos of real-time one-shot pose estimation, we refer the reader's attention to the accompanying video.

\subsubsection{Attention maps for object detection} \label{sec:object_crop}
In our experiments on the CO3D dataset, object detection is an upstream process, and we operate on images closely cropped to the objects of interest. We show that we are able to achieve the same accurate object detection and cropping in a fully autonomous setting (running pose estimation on a robot arm at 15Hz) through running ViT inference on each image twice. This process is shown in \cref{fig:bbox_proc}. 

In the first pass, the whole field-of-view image from the wrist-mounted RealSense camera is processed. A threshold (0.05) on the attention map from the first pass is used to produce a binary segmentation mask, from which the largest connected component is taken to be the object of interest. The resulting bounding box is up-sampled to the original image size, and expanded by 10\% so as to be sure to capture the whole object. This box is used to produce a closely cropped image whose ViT features are used for correspondences and pose estimation. 

\subsubsection{Bounding boxes for visualising pose estimates} \label{sec:bbox}

\begin{figure}[htb!]
\centering
  \includegraphics[width=\columnwidth]{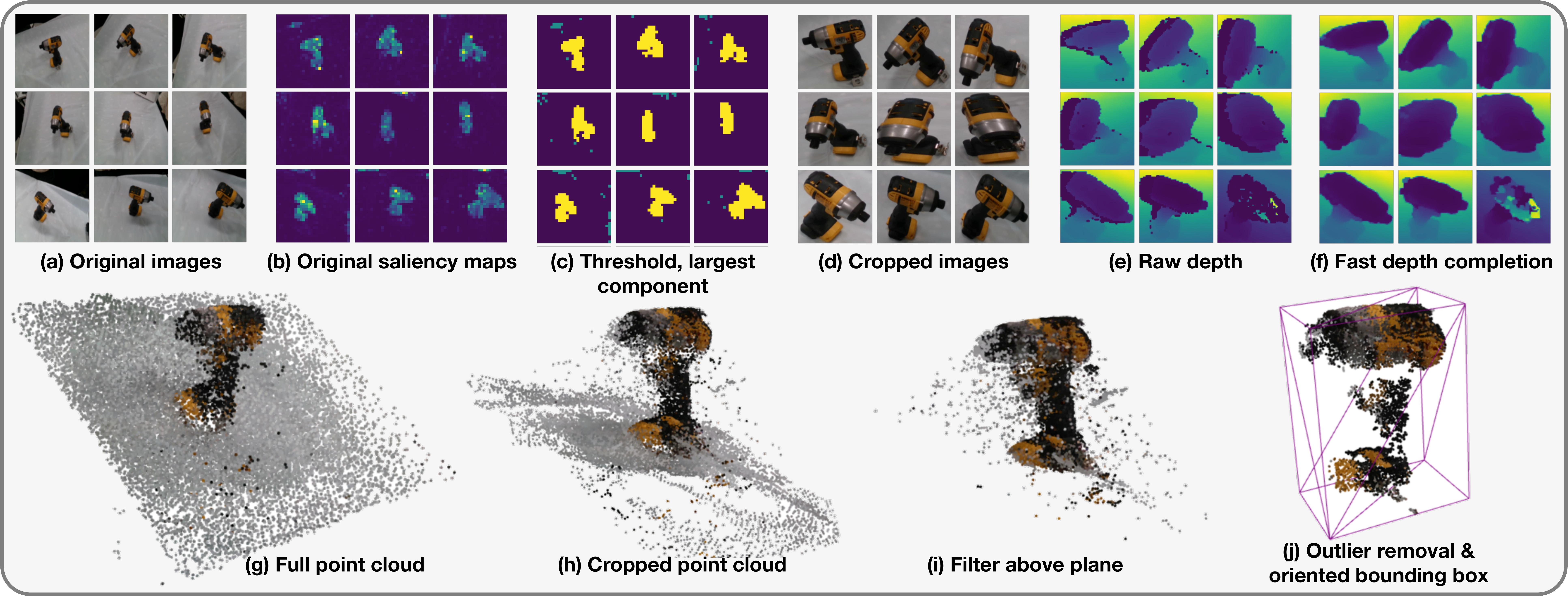}
  \caption{Point cloud processing for fitting oriented bounding boxes to reference objects. \textbf{(a)} Original images from RealSense camera, 720x1280 resolution centre-cropped to 720x900 then resized to 224x224 for ViT processing. \textbf{(b)} Saliency aggregated from the ViT attention maps. \textbf{(c)} A threshold of $>0.05$ on saliency produces a binary foreground mask - here, the largest connected component is shown in yellow (other `foreground' in blue). \textbf{(d)} Images cropped to the box described by the largest connected component. \textbf{(e)} Raw depth from the RealSense D435i, following same crop. \textbf{(f)} Depth maps following fast inpainting process. \textbf{(g)} Point cloud from backprojecting original images. \textbf{(h)} Point cloud using saliency-based cropping. \textbf{(i)} Point cloud following plane detection. \textbf{(j)} Final point cloud following outlier removal with a KNN criteria, and oriented bounding box fitting.}
  \label{fig:bbox_proc}
\end{figure}

Although we describe our method as being \textit{one-shot} pose estimation because of its use of a reference sequence to describe a target object category, we design our system to require no manual labelling or manual processing of this reference object, such that the whole method could in practice be deployed as fully autonomous. 

Our method estimates a 7 DoF rigid body transforming $(\hat{\lambda}, \hat{\mathbf{R}}, \hat{\mathbf{t}})$ between a category's reference object, and a target object, as described in \cref{sec:umeyama}, where 6D pose is given by SO(3) rotation $\hat{\mathbf{R}}$ and translation $\hat{\mathbf{t}}$. While this captures object pose from a mathematical point of view, it does not immediately offer a way of \textit{visualising} the pose estimates. To facilitate this in a fully automated pipeline, we fit an oriented bounding box to the \textit{reference} object for a category, and transform this by the estimated 7 DoF rigid body transform to visualise pose estimates for the target objects.

The process of fitting an oriented bounding box to a captured reference object is shown in \cref{fig:bbox_proc}. The first stages of close cropping are described in \cref{sec:object_crop}. Subsequently, masking and point cloud outlier removal steps produce a filtered point cloud, to which an oriented bounding box is fitted based on the PCA of the convex hull of the filtered point cloud. This is a fast approximation to the minimum-volume bounding box, which tends to produce intuitive, easily interpreted bounding boxes around object volumes \cite{Wu2022}. We use Python bindings to the Open3D library to perform this final stage \cite{Zhou2018open3d}.

For outlier removal, we use a fast K-nearest neighbours based approach using Pytorch3D \cite{Pytorch3d}. For an inlier point, we require that its 10 nearest neighbours be within \SI{5e-5}{m}.

\end{document}